\RequirePackage{fix-cm}
\documentclass[smallcondensed]{svjour3}  
\smartqed 
\usepackage{graphicx}
\usepackage[numbers]{natbib}
\usepackage{float}
\usepackage[title]{appendix}
\usepackage{subcaption}
\usepackage{amsmath}
\usepackage{graphicx}
\usepackage{soul,color}
\soulregister\citep7
\soulregister\ref7
\soulregister\pageref7
\usepackage{multirow}
\usepackage{graphicx}
\usepackage[table,xcdraw]{xcolor}

\usepackage{algorithm}
\usepackage{algpseudocode}

\journalname{Applied Intelligence}
\begin{document}

\title{Forecasting the abnormal events at well drilling with machine learning}
\author{Ekaterina Gurina         \and
        Nikita Klyuchnikov \and
        Ksenia Antipova \and
        Dmitry Koroteev
}

\institute{E.Gurina \at
              Skolkovo Institute of Science and Technology (Skoltech), 121205, Moscow, Russia \\
              Digital Petroleum, Skolkovo Innovation Center, 121205,  Moscow,  Russia
              \email{Ekaterina.Gurina@skoltech.ru}           
           \and
           N. Klyuchnikov \at
              Skolkovo Institute of Science and Technology (Skoltech), 121205, Moscow, Russia \\
              Digital Petroleum, Skolkovo Innovation Center, 121205,  Moscow,  Russia
           \and
            K. Antipova \at
              Skolkovo Institute of Science and Technology (Skoltech), 121205, Moscow, Russia \\
              Digital Petroleum, Skolkovo Innovation Center, 121205,  Moscow,  Russia
         \and
            D. Koroteev \at
              Skolkovo Institute of Science and Technology (Skoltech), 121205, Moscow, Russia \\
              Digital Petroleum, Skolkovo Innovation Center, 121205,  Moscow,  Russia
}
\date{Accepted: 14 November 2021, Published 10 January 2022}
\maketitle

\begin{abstract}
We present a data-driven and physics-informed algorithm for drilling accident forecasting. The core machine-learning algorithm uses the data from the drilling telemetry representing the time-series. We have developed a Bag-of-features representation of the time series that enables the algorithm to predict the probabilities of six types of drilling accidents in real-time. The machine-learning model is trained on the 125 past drilling accidents from 100 different Russian oil and gas wells. Validation shows that the model can forecast 70\% of drilling accidents with a false positive rate equals to 40\%. The model addresses partial prevention of the drilling accidents at the well construction.
\keywords{Bag-of-features \and Directional drilling \and  Machine learning \and Classification \and Telemetry}
\end{abstract}

\section{Introduction}
Drilling of oil and gas wells is usually accompanied by uncertainty about geological conditions. Regardless of the available information, unusual situations inevitably happen during the drilling process, the most unpleasant of which are drilling accidents (emergencies that make any further work impossible or delay future activity).

Engineers using surface telemetry usually detect most of the accidents. Noticing the non-standard behavior of the main drilling parameters at mud log data, a qualified engineer may forecast the accident. In most cases, such behavior or pattern signalizes not only the possibility of an accident but also about a particular type of accident, and, consequently, by detecting pre-accident non-standard behavior, most accidents can be predicted. Since most of the engineers support a large number of wells simultaneously, they usually do not have time to monitor all the wells online, and thus they can miss the patterns. Developing a system that can forecast accidents using real-time surface telemetry data and artificial intelligence techniques is one of the possible solutions for such a problem. AI solutions were successfully applied in different areas of oil and gas industry \citep{hajizadeh2019machine, bangert2021machine}, mining sector \citep{ali2020artificial}, medicine \citep{samhitha2020improving}. For example, in papers \citep{kumar2021recent, IT_article} authors pointed out several business effects such as reduction of fraud operation in bank area from five to one percentage, increasing the general customer satisfaction and speed during online operations in two times. 

Nowadays, there are few existing solutions for the drilling accidents forecasting problem. For example, in papers \citep{sadlier2013automated, ferreira2015automated} authors developed a system that can compare the real-time situation at drilling rig with similar situations from a database of previously drilled wells with similar configurations. However, such a solution is not generalizable from field to field since there might not be a similar situation inside the database.

Another solution for accident detection while drilling was applied and described in paper \citep{gurina2020application}. This approach was based on summary statistics (mean, standard deviation, slope, and ratios) of each interval as features. However, it does not show the high values of metrics during the accident forecasting \citep{antipova2019data}. The described approach can be used as the model for the state-of-the-art comparison for our solution.

In papers, \citep{abu2010application, abu2015automated} researchers used simulated data while drilling from a mud log station as input features to the clustering algorithm with further prediction based on neural network. As output variable, they assign two possible states: common situation and pre-accident situation. This approach does not allow for assessment type of accident and might show bad performance while testing real-time data. 

A similar approach used mud telemetry data as input data and neural network as a classifier as described in papers \citep{aljubran2021deep, qodirov2020development}, which allows forecasting sticking and loss of circulation accidents while drilling. A similar solution can also be applied for benchmarking our approach. 

The summary of the described solutions is presented in Table \ref{tab:comparision}. Based on research results obtained through summarizing and analyzing existing solutions, we conclude that it is necessary to improve predictive methods considering the above disadvantages, using techniques not applied before for accident forecasting problems.

\begin{table}[]
\centering
\caption{Comparision of existing solutions for the drilling accidents forecasting problem.}
\label{tab:comparision}
\resizebox{\textwidth}{!}{%
\begin{tabular}{|c|c|c|c|c|}
\hline
\textbf{№} & \textbf{Idea} & \textbf{Proposed solution} & \textbf{Drawbacks} & \textbf{Papers} \\ \hline
1 & \begin{tabular}[c]{@{}c@{}}Compare the real-time \\ situation at drilling rig \\ with similar situations \\ from a database \\ of previously \\ drilled wells with \\ similar configurations\end{tabular} & \begin{tabular}[c]{@{}c@{}}Multivariate statistical\\  algorithms and similarity\\  learning\end{tabular} & \begin{tabular}[c]{@{}c@{}}Not generalizable \\ from field to field, \\ requires large database \\ of similar situations\end{tabular} & \citep{sadlier2013automated, ferreira2015automated} \\ \hline
2 & \begin{tabular}[c]{@{}c@{}}Identify the accidents \\ of different types based on \\ the summary statistics \\ for mud telemetry features \\ and ML classifier\end{tabular} & \begin{tabular}[c]{@{}c@{}}Summary statistics \\ of each interval \\ as features and \\ gradient boosting \\ classifier\end{tabular} & \begin{tabular}[c]{@{}c@{}}Not designed for forecasts, \\ therefore, underperforms approaches\\  aimed at forecasts\end{tabular} & \citep{gurina2020application, antipova2019data} \\ \hline
3 & \begin{tabular}[c]{@{}c@{}}Identify if the considering \\ interval is normal\\  based on features \\ defined from clustering \\ algorithm and simulated data\end{tabular} & \begin{tabular}[c]{@{}c@{}}Clustering algorithm\\  with \\ further NN \\ classifier\end{tabular} & \begin{tabular}[c]{@{}c@{}}Does not allow to assess \\ type of accident\end{tabular} & \citep{abu2010application, abu2015automated} \\ \hline
4 & \begin{tabular}[c]{@{}c@{}}Use mud telemetry data \\ as input data to neural network \\ to forecast \\ sticking and loss \\ of circulation\end{tabular} & \begin{tabular}[c]{@{}c@{}}NN classifier based \\ on raw mud telemetry \\ data obtained \\ with sliding window\end{tabular} & \begin{tabular}[c]{@{}c@{}}No statistically \\ significant results\end{tabular} & \citep{aljubran2021deep, qodirov2020development} \\ \hline
\end{tabular}%
}
\end{table}

This paper aims to employ pattern-mining techniques for application to accident forecasting at well drilling. Our predictive model detects pre-accident patterns that can be frequently observed within six hours before the accident and rarely occur during the normal operating mode. This paper’s main novelty is applying the Bag-of-features method together with wavelet representation of multivariate time-series data to forecast drilling accidents of six possible types: stuck, washouts of drilling pipes, mud loss, breaks of drilling pipes, fluid shows, and packing. 

The main paper contributions are the following:
\begin{itemize}
    \item We collected a large dataset of mud telemetry data for wells in the oilfield in Russia and overviewed data quality issues.
    \item  We developed a system that uses multivariate time-series data and transforms it into the Bag-of-features representation, using the clustering techniques and wavelet transform.
    \item Based on the obtained representation, we solved the multi-label machine learning problem, compared results with a random baseline, and benchmarked it with several state-of-the-art methods for drilling accidents forecasting. We assessed the obtained solution in terms of ROC AUC metric value \citep{hanley1982meaning} and reflected it in value for business.
    \item Additionally, we made an ablation study of the proposed solution together with model fine-tuning and performance optimization.
\end{itemize}

The paper is organized as follows: Section \ref{sec: state_of_art} contains a brief introduction to the articles related to existing solutions for the drilling accidents forecasting problem together with the articled related to the time-series classification tasks. In Section \ref{sec:methodology} we presented the methodology adopted for the current study. Section \ref{sec:results} provides a detailed analysis of the proposed solution and a set of recommendations to increase the model quality and interpretability. Finally, Sect. \ref{sec:conclusion} presents the conclusions of this study.

\section{Literature review}
\label{sec: state_of_art}
Pre-accident patterns that can be frequently observed within six hours before the accident and rarely occur during the normal operating mode in most cases can be distinguished on telemetry logs. As a pre-accident pattern for a stuck accident, a drilling engineer may observe several overpulls and drags down of the drilling strings, characterized by the rapid increase or decrease of a hook load over its weight of pipes. At the same time, if the drilling column was already stucked, one may observe no column movement. In the case of washouts of the drill string, a decrease in pressure at a constant flow rate might be observed (\citep{ Washouts}). A similar pattern might be noticed in the case of mud loss accidents, where a pressure decrease is observed together with a decrease in the volume in the tanks. Both breaks of drilling pipes and packing are characterized by the "jumps" of the torque moment signals. However, there is a sharp drop in weight after the accident moments in the case of breaks of drilling pipes. Fluid shows can usually be forecasted by the small increase of the mud output volume and small increases of the gas level. So, both pre-accidents and accidents patterns can be described as a set of anomaly behavior in time-series signals for the last period. 

Based on the physics of the pre-accidents and accidents patterns, the problem of drilling accident forecasting based on mud logs can be formulated as follows: given the multivariate time series database (mud logs), one should build a model that for different types of drilling accidents will detect non-standard drilling behavior that usually occurs only before the accident, and rarely in case of normal operations. 

Considering the unsupervised approach,  one could solve a problem based on a distance comparison of time series. In papers \citep{li2020adaptively, serra2014empirical}, authors suggest dividing the entire database of time series into several clusters, using, for example, Euclidean or the Dynamic Time Wrapping distances. Since most of the logs represent high noise and frequent data, the proposed division might be invalid and time-consuming.  At the same time, the described approach also shows a low ability to handle the missing values, which are usually presented in telemetry data. 

Another group of unsupervised approaches is related to the usage of recurrent neural nets. In papers \citep{malhotra2017timenet, sagheer2019unsupervised, yoon2019time}, authors train the recurrent neural network or generative adversarial networks that reconstructs the input time series from sequence-to-sequence learned representation of the initial time series. However, these methods are not suited to long time series due to the sequential nature of a recurrent network, while the length of anomaly intervals may vary from $10$ to $300$ minutes long. 

The problem of accident forecasting also can be considered as a problem of time series classification. Given the multivariate time series dataset, the machine learning algorithm has to find a function that is as close as possible to the correct classification (whether the drilling accident will happen in a short time or not). It is necessary to find a time series representation that can be used as an input value for the machine-learning algorithm and be as informative as possible for accident forecasting. Since there are many time-series representation methods, we will show only a few of them relevant to our problem. The full overview of possible time series representations can be found in papers \citep{abanda2019review, fu2011review, fawaz2019deep}.

One technique used nowadays is a representation of the time series by their distance from the initial time series to the \textit{k} best shapelets, obtained from the database \citep{bostrom2017binary, rakthanmanon2013fast}. According to \citep{grabocka2014learning}, shapelets are time series subsequences that are identified as representative objects of a particular class. Shapelets are not sensitive to high noised and frequent data as telemetry logs. They can also be easily interpretable, which may give us explanatory insights into possible precursor patterns. However, shapelets cannot change their scalability, while pre-accidents' patterns can differ by their length. Besides, shapelets also require an extensive search for the \textit{k} best shapelets forms from a large space of accident precursors and drilling logs with normal behavior. 

The next group of authors postulates the use of time series transforms named Bag-of-features \citep{Wang2016RPMRP, hatami2019bag, begum2014rare}. The current approach is mostly used in natural language processing, where each word in the sequence is labeled by the particular symbol or index, named codeword, from the defined codebook. In our case, instead of words, we should label small segments of time series using clustering algorithms. Such procedure is similar to the shapelets forming when the particular cluster stands for a certain short time series pattern. Considering interval, which includes the labeled segments, one may build a histogram of obtained labels further used as input features for the machine learning algorithm. In paper \citep{wang2013bag}, the Bag-of-features approach was successfully applied to biomedical time series classification, however, the results were obtained only for univariate time series. Compared to the other representations, Bag-of-features can overcome the disadvantages of previous methods described above, such as scalability of the patterns included in the model and learning long time-series patterns. In contrast, the method's drawbacks are the computational speed and high feature space dimension, limiting its application for large datasets.

Several authors use neural networks, namely convolution networks, both for feature extraction and classification. For example, in papers, \citep{zheng2014time, wang2017time, karim2019multivariate} authors use the Multi-Channels Deep Convolution Neural Networks model, where each channel takes a single dimension of multivariate time series as input and learns features individually by passing values through convolution layers.  In most cases, working with neural networks implies a large sample size. In our case, the training set consists of $22000$ time series segments one hour long, obtained on $100$ wells, which might be sufficient for the neural network training.


\section{Methodology}
\label{sec:methodology}

\subsection{Data overview}
Since the accident forecasting model's purpose is to apply extra support to the decision-making process, this model should use the available parameters in most wells used by the engineer to validate the drilling process. 

\begin{figure}[!ht]
     \centering
     \includegraphics[width=\textwidth]{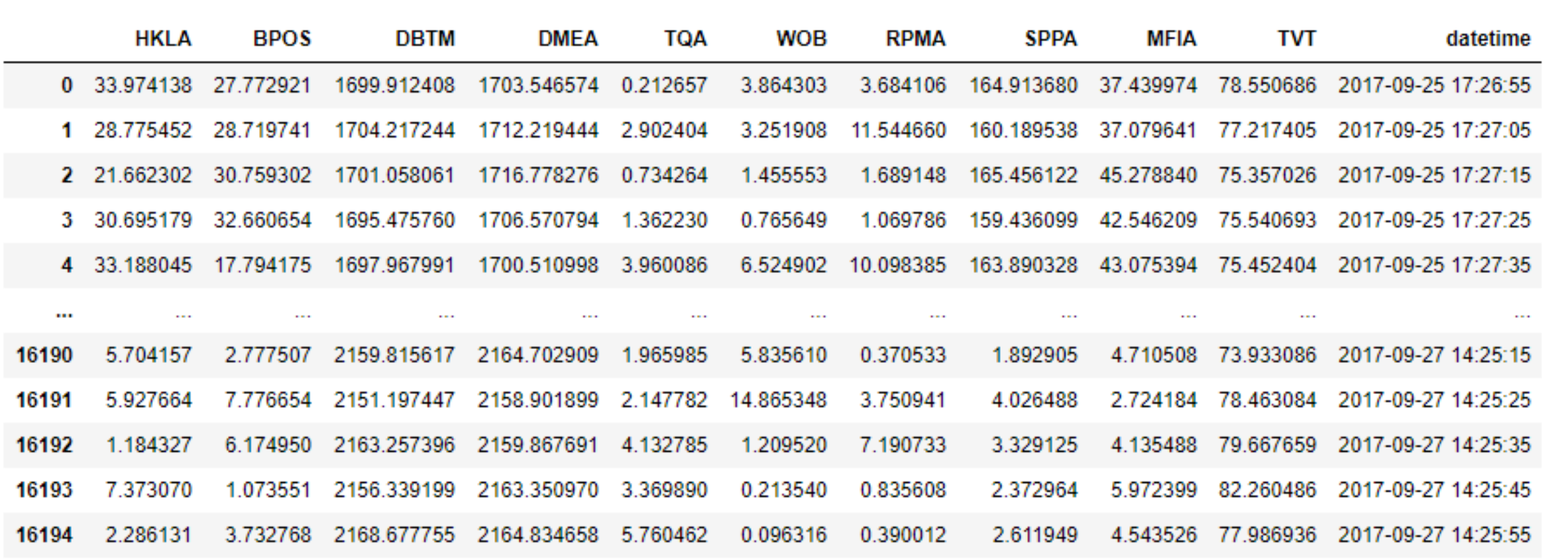}
     \caption{Telemetry log records for the Well 1. The units for each corresponding parameter are shown in the Table \ref{tab:MWD_params}.}.
     \label{fig:df_ex}
\end{figure}

Nowadays, most of the drilling rigs are equipped with automated stations, named mud telemetry systems, for the drilling process control in real-time. Such stations are operated directly and solve several geological and technological tasks aimed at successfully drilling the well. Transmitting the telemetry parameters recorded online from different sensors both from the drill bit and rig station to the drilling engineers,  one may assess the stability of the drilling process. Mud telemetry data report the measurements of parameters related to the drilling process sampled every 5 seconds. The description of such parameters, units, and possible value ranges is shown in Table \ref{tab:MWD_params}. If the considered time interval included abnormal values, they were replaced by the last valid value. An example of the data containing the telemetry log records for the particular well is presented in Figure \ref{fig:df_ex}. All wells were drilled in different regions and had vertical and horizontal sections. The training data consist of different segments that correspond to such drilling operations as a trip in, trip out, drilling, cleaning, and reaming.

\begin{table}[!ht]
\centering
\caption{Telemetry logs parameters used for the accident forecasting model}
\begin{tabular}{|c|c|c|c|c|c|}
\hline
 & Parameter & Units & \begin{tabular}[c]{@{}c@{}}Minimum \\ value\end{tabular} & \begin{tabular}[c]{@{}c@{}}Maximum\\ value\end{tabular} & Abbreviation \\ \hline
1 & Hook load & Tons & 0 & 300 & HKLA \\ \hline
2 & Block Position & Meter & 0 & 50 & BPOS \\ \hline
3 & Depth Bit (meas) & Meter & 0 & 10000 & DBTM \\ \hline
4 & Depth Hole (meas) & Meter & 0 & 10000 & DMEA \\ \hline
5 & Rotary Torque & kN*m & 0 & 140 & TQA \\ \hline
6 & Weight-on-Bit & Tons & 0 & 25 & WOB \\ \hline
7 & Rotary Speed & rot.per min. & 0 & 200 & RPMA \\ \hline
8 & Standpipe Pressure & atm & 0 & 350 & SPPA \\ \hline
9 & Mud Flow In & l/s & 0 & 65 & MFIA \\ \hline
10 & Volume in tanks & m3 & 0 & 240 & TVT \\ \hline
11 & Gas content & \% & 0 & 1 & GASA \\ \hline
\end{tabular}
\label{tab:MWD_params}
\end{table}

Considering the collected database of drilling accidents, it consists of 125 cases of 6 different types (Table \ref{tab:accident_types}), which happened in $100$ different Russian oil and gas wells in North and West Siberia. Definition of each accident type can be found in \citep{glossary}. Such accidents were chosen by the number of available cases and the possibility of being distinguished visually on telemetry logs. 

\begin{table}[!ht]
\centering
\caption{Drilling accident types, presented in the database}
\begin{tabular}{|c|c|c|}
\hline
 & Accident type & Number of cases \\ \hline
1 & Stuck & 55 \\ \hline
2 & Washouts of drilling pipe & 20 \\ \hline
3 & Mud loss & 16 \\ \hline
4 & Breaks of drilling pipe & 15 \\ \hline
5 & Fluid shows & 10 \\ \hline
6 & Packing & 9 \\ \hline
\end{tabular}
\label{tab:accident_types}
\end{table}

\begin{figure}[!ht]
     \centering
     \includegraphics[width=0.6\textwidth]{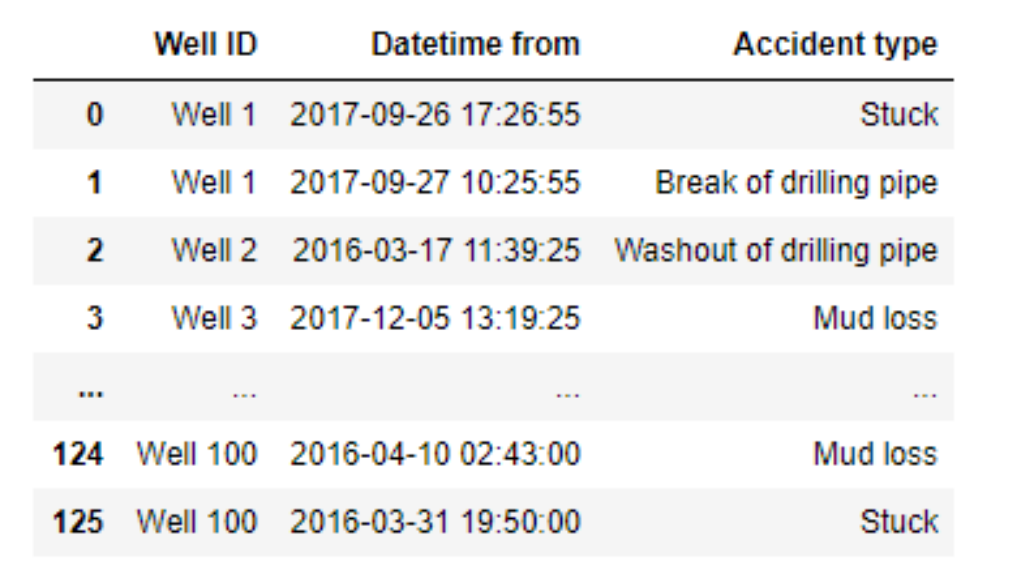}
     \caption{Overview of reference data collected for the accident forecasting problem. One row in the reference table relates to the particular well, accident type, and time when the accident occurred.}
     \label{fig:ref_ex}
\end{figure}

An expert analyzed each of the telemetry logs for the presence of anomaly behavior and why the particular accident occurred. For all cases, the expert uses only the telemetry parameters, presented in Table \ref{tab:MWD_params}, and highlights the corresponding interval if abnormal behavior can be clearly distinguished in telemetry log data. An example of the expert reference table for all wells is given in Figure \ref{fig:ref_ex}.

\subsection{General scheme of the model}
Based on methods considered in Section \ref{sec: state_of_art}, the problem of drilling accident forecasting can be solved as a supervised machine learning problem trained on time-series database representation via the Bag-of-features approach. Let us consider two main steps of the model: feature generation and classification problem. 

\subsubsection{Feature generation}
\label{sec:feature_generation}
According to the Bag-of-features approach, we should assign one label from the codebook to the time series segment. To obtain the codebook of possible labels, we decided to train the clustering algorithm to assign the cluster number to the small segments of the time series, named $\tau$ segments. We also decided to represent the $\tau$ segment in the frequency domain using the wavelet transform since it is better represents the shape of the time series and better preserves similarity during shifting and stretching deformations. Since we mostly deal with non-stationary signals, Discrete Wavelet transform was chosen as the most promising one for such representation \citep{milon2017comparison, sifuzzaman2009application}.  The detailed description of the Discrete Wavelet transform might be found in paper \citep{tzanetakis2001audio}, while the particular type of wavelet and level of decomposition used for the current problem will be discussed in section \ref{sec:results}.

\begin{figure}[!ht]
    \centering
    \includegraphics[width = 0.95\linewidth]{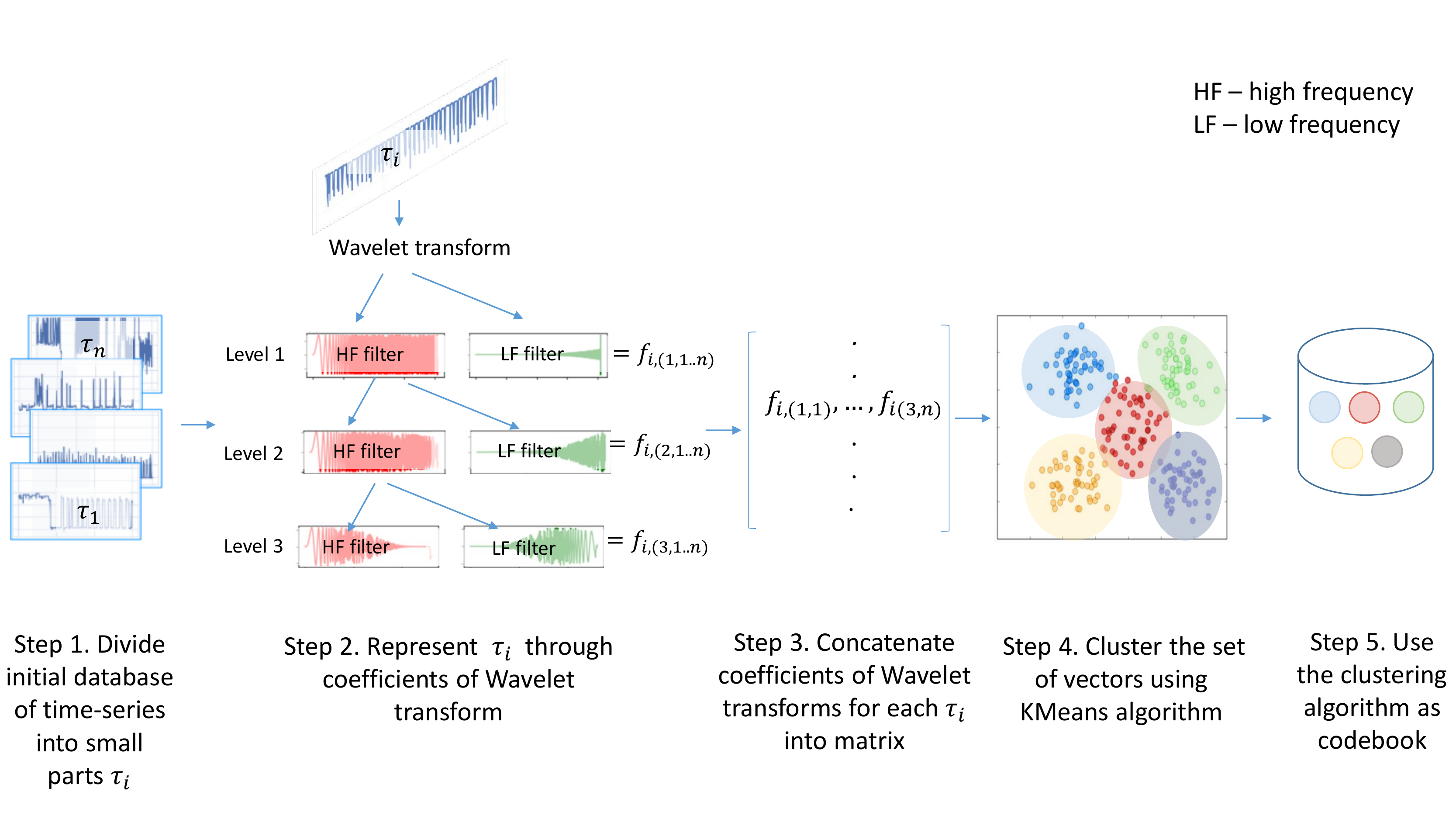}
    \caption{The procedure of Bag-of-features representation. Similar to the \citep{wang2013bag}, the codebook is constructed by clustering small segments from training data. Circles of different colors stand for codewords in the codebook, i.e., cluster of time series. The procedure was carried out for each telemetry parameter separately.
    } 
    \label{fig:tau_learning}
\end{figure}

The scheme of the codebook construction process is shown in Figure \ref{fig:tau_learning} \and presented as pseudo-code in algorithm \ref{alg:codebook}. The represented procedure was carried out separately for each of the drilling parameters in the following way:
\begin{enumerate}
    \item For each log presented in the database: split it into small parts $\tau_i$ of length $\Tilde{\tau}$, using sliding window technique with step $q$;
    \item Represent $\tau_i$, using coefficients from Wavelet transform;
    \item Concatenate coefficients of wavelet transforms for each $\tau_i$ into the feature-matrix of $\tau$-parts;
    \item Cluster the set of feature-vectors, obtained on the previous step, into $K$ clusters, using the k-means algorithm
    \item Use the trained clustering algorithm as a codebook.
\end{enumerate}

Before the k-means procedure, we did not scale the set of feature-vectors since the scales of input features do not differ a lot (no more than ten times). Additionally, several variations of k-means algorithms were tried: usual k-means, k-means with different weights, k-means with assigning several clusters for one input sample. However, they did not bring significant differences in clustering quality metrics.

\begin{algorithm}
\caption{An algorithm for codebook construction}
\label{alg:codebook}
\begin{algorithmic}
\Require{Set of time-series - $X$, Telemetry feature for codebook construction - $y$}
\State {matrix =[ ]}
\For{x in $X$}
    \State {$tau_{i}$ = SplitTimeSeries(x[y])}
    \For{$tau_{ij}$ in $tau_{i}$}
    \State {coefficients = GetWaveletTransform($tau_{ij}$)}
    \State {coefficients $\rightarrow$ matrix }
    \EndFor
  \EndFor
\State{codebook = KMeans(matrix)}
\end{algorithmic}
\end{algorithm}

By obtaining the labels for all drilling parameters for $\tau$ segments, one could build a histogram of labels, predicted for the time series intervals, named $t$ segments, using the following steps: 
\begin{enumerate}
    \item For each of the telemetry parameters, using sliding window technique with step $h$, obtain the set of segments $t_j$ of length $\Tilde{t}$ for the initial time series $T$;
    \item Divide each segment $t_j$ into small parts $\tau_i$; 
    \item Assigning the label for each of the $ \tau_i$ with trained KMeans algorithm, obtain the set of clusters labels for the particular $t_j$;
    \item Obtain the vector of features for the particular telemetry parameter and segment $t_j$, building a histogram of labels distributions ;
    \item Combine the vectors for each telemetry parameter into one vector of features related to the segment $t_j$.
\end{enumerate}

\begin{figure}[!ht]
    \centering
    \includegraphics[width = 0.9\linewidth]{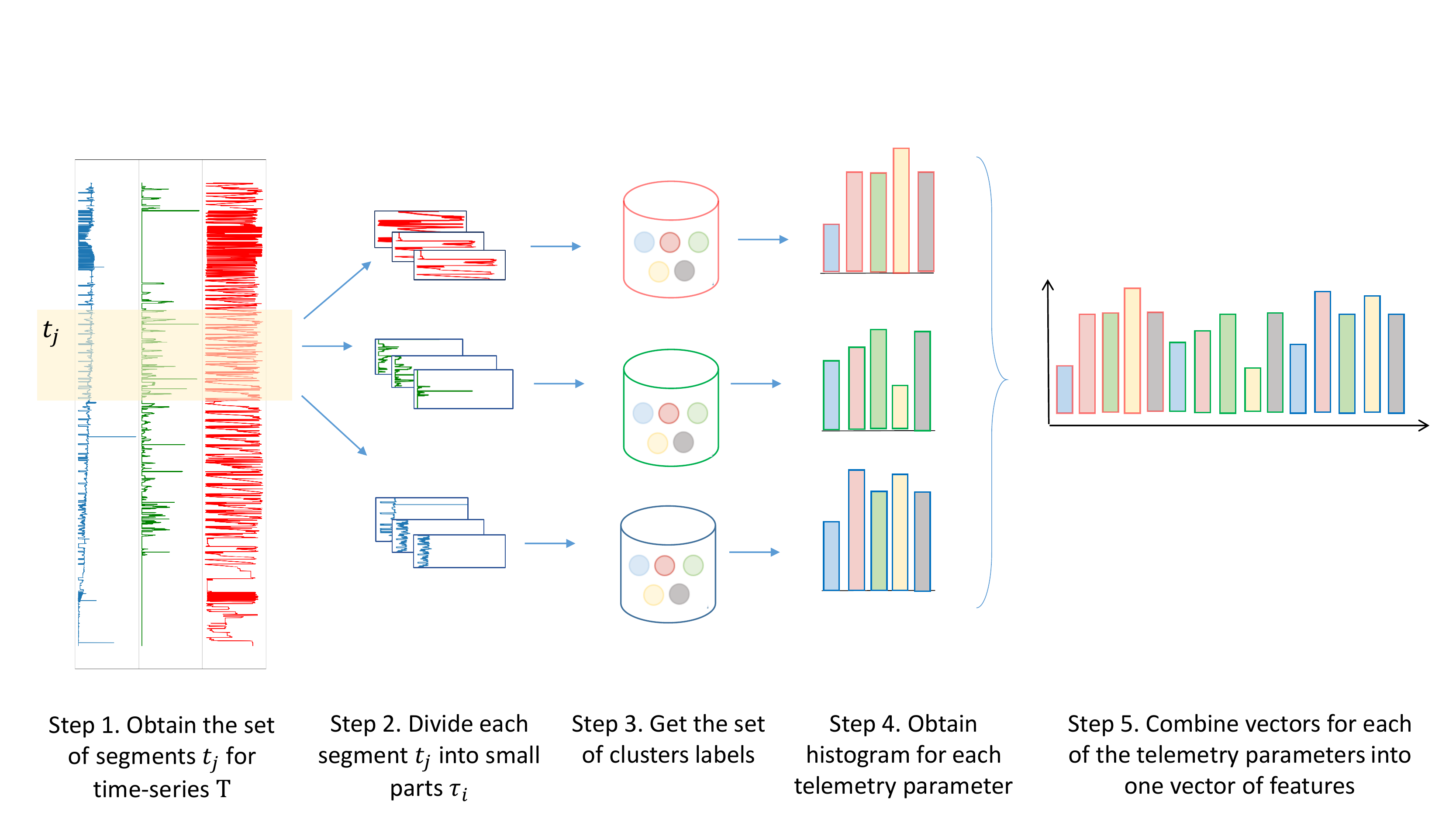}
    \caption{Feature generation scheme. Using the Bag-of-features approach, one could build a histogram of labels for each telemetry parameter. The final vector of features for the segment $t_j$ was obtained by concatenating several histograms into one vector.} 
    \label{fig:hist_gen}
\end{figure}

The scheme of the feature generation procedure is shown in Figure \ref{fig:hist_gen} and presented as pseudo-code in algorithm \ref{alg:hist}. An example of the generated histogram for the particular part of telemetry log data is presented Appendix \ref{Appendix_example}. Using described strategy, one could obtain time-series database representation via the Bag-of-features approach and build the matrix of features for further experiments' training and validation phases.   

\begin{algorithm}[!ht]
\caption{An algorithm for histogram generation during feature engineering.}
\label{alg:hist}
\begin{algorithmic}
\Require{Data - $X$, Set of telemetry features - $Z$}
\State {histograms = [ ]}
\For{feature in $Z$}
    \State {$t$ = SplitTimeSeries(X[feature])} 
    \State {$\tau$ = [ ]}
    \For{$t_j$ in t}
        \State {$\tau_{ij}$ = SplitTimeSeries($t_j$)}
        \State {$\tau_{ij}$ $\rightarrow$ $\tau$ }
        \EndFor
    \State{labels = codebook($tau$)} 
    \State{historgam = BuildHistorgram($tau$)} 
    \State {historgam} $\rightarrow$ \texttt{histograms}
  \EndFor
\State {feature matrix = Concatenate(histograms)}
\end{algorithmic}
\end{algorithm}

\subsubsection{Classification problem}
Since the model should forecast the probability of several types of drilling accidents, we use the multitarget strategy, where a single classifier is trained per class with the samples of that class as positive ones and all other samples as negatives. We decided to use a model based on the Gradient Boosting of decision trees for a single classifier. They are relatively undemanding in terms of sample size and data quality and learn quickly with a large number of features, which was also shown in papers \citep{Grad_boost, Grad_boost_2}. In our case, we use Gradient Boosting algorithm with 50 estimators, learning rate equals 0.05, max depth of 10, while the subsample and colsample by tree correspond to 90\%. In addition, we increased the weight of the positive class five times. Since we have an imbalanced dataset, we also tried under-sampling and oversampling techniques, however, none significantly improved model quality.

The training process uses mud telemetry intervals of $24$ hours long before the accident. As input values, the model uses features of selected intervals obtained during the feature generation phase (Section \ref{sec:feature_generation}). During the model's training,  we assume that considered interval leads to drilling accidents if it was six hours before the beginning of the accident. Since the model's purpose is to forecast the accident of all six possible types, we would not separate on different cases (e.g., mud losses can develop over time while pipe breaking may be a sudden event). Besides, if considered interval leads to several types of drilling accidents, a positive target value is assigned for all selected types. Since pre-accidents patterns differ by length, such six-hour intervals may contain parts of normal drilling behavior, marked as pre-accident ones. However, we will assume that their influence during model training will not be significant since the training sample will contain many similar normal intervals sampled from the rest of the training set, diluting the influence of normal intervals annotated as pre-accident. 

During the real-time model performance, at each time moment, the model generates features for the last received part of data of length $t$ and returns the probability of whether the input data contains the pre-accident patterns for each of the accident's types. Described procedure presented as pseudo-code in algorithm \ref{alg:test}. The model can be switched on with different time steps, however, the model results will correspond only to when the model was switched on.

\begin{algorithm}
\caption{An algorithm for a testing model for drilling accident forecasting.}
\label{alg:test}
\begin{algorithmic}
\Require{Data of one hour length - $X$}
\State{features = GetHistorgram(X)}
\State{probabilities = PredictProbabilities(features)}
\If{any(probabilities) $\geq$ threshold}
\State{Alarm user}
\EndIf
\end{algorithmic}
\end{algorithm}

\subsection{Validation procedures}
\label{sec:val_proc}
To validate model performance, we build the Receiver operating characteristic (ROC) curve and calculate the area under it (ROC AUC metric). Since we deal with a multiclass classification problem, the current metric was calculated for each instance and then averaged. For the accident forecasting metrics, we assume that the accident was forecasted if the model alarm happened before the $6$ hours of the accident and the type was predicted correctly. Otherwise, the model warning was counted as a false alarm. 

\begin{figure}[!ht]
    \centering
    \includegraphics[width = 1\linewidth]{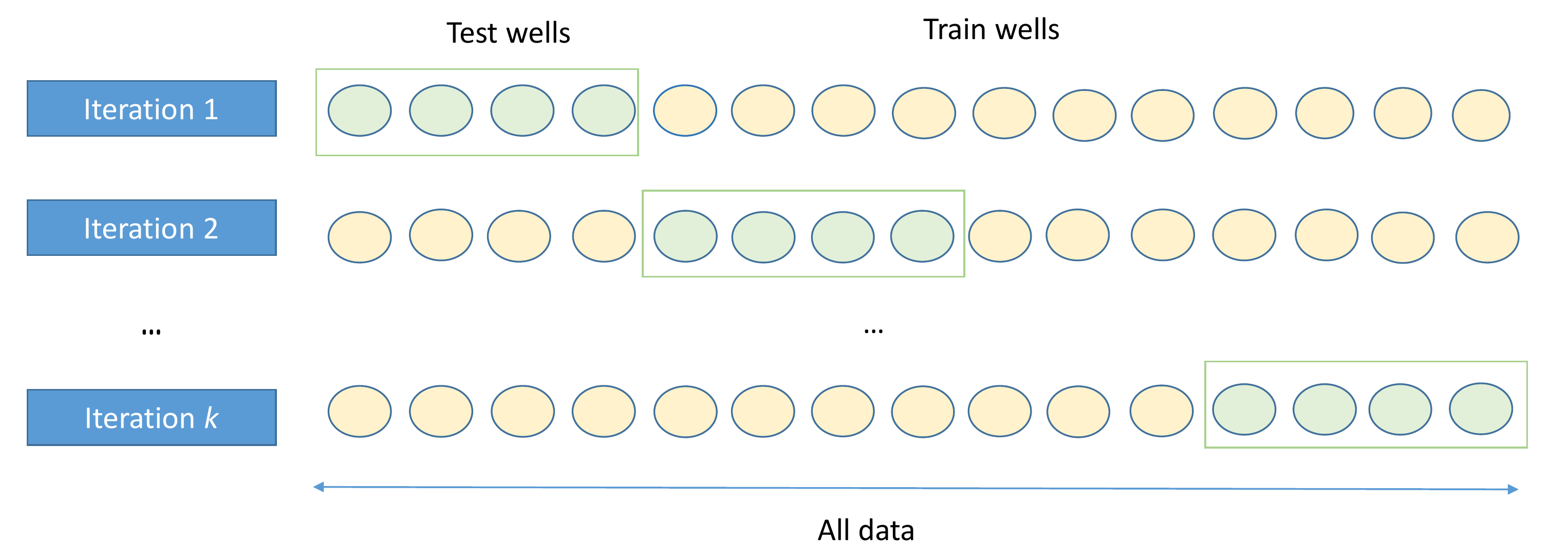}
    \caption{Scheme of $k$-fold cross-validation procedure. For each iteration, one could generate two sets of non-intersecting wells used as test and train samples. One could get the model's final metric value by obtaining the test sample results on each iteration.
    }
    \label{fig:cross-val}
\end{figure}

Metrics were calculated using a $5$-fold cross-validation technique \citep{Foundation_of_ML}. Scheme of cross-validation procedure is presented in Figure \ref{fig:cross-val}. Generating two random sets of non-intersecting wells, one could obtain train and test samples, picking the wells from the corresponding set. On each step, the train set always contains examples of all six accident types. The probability that all anomaly intervals of the same accident type will not appear in the training sample for some fold is less than 0.01 percent. The model was trained on the wells, which indices were chosen as train one. Validation carried out on the test set, which for each well includes the $24$ hour intervals before the accident and $20$ intervals of eight hours length of normal drilling behavior. Repeating the described procedure $k$ times, one could obtain the final ROC AUC score.

\section{Results and discussions}
\label{sec:results}
\subsection{Selection of the main hyperparameters}
\begin{figure}[!ht]
    \centering
    \includegraphics[width = 1\linewidth]{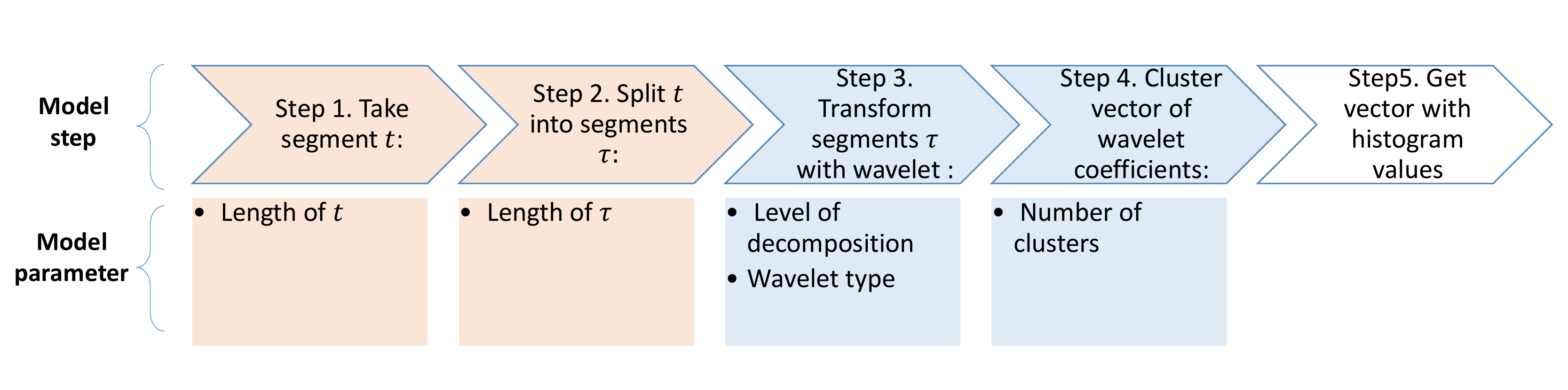}
    \caption{Scheme of the main hyperparameters versus model step. For each model step, one could see related hyperparameters of the model that need to be selected. Blue and orange colors stand for two groups of parameters that were selected at the same time.} 
    \label{fig:scheme_hyper}
\end{figure}

Before introducing the general model quality metrics, we conduct several experiments to select the primary model hyperparameters. The main hyperparameters' scheme versus model steps is shown in Figure \ref{fig:scheme_hyper}. We do not show the KMeans algorithm's parameters on the scheme since tuning these parameters is carried out with common procedures, e.g., described in \citep{claesen2015hyperparameter, bergstra2013making}. 
According to Figure \ref{fig:scheme_hyper}, the drilling forecasting model has two main groups of parameters (blue and orange boxes), where parameters marked with one color mean that their selection took place within the same experiment. 

Based on the experiment setup, one could notice that the true optimal combination may be different since all hyperparameters influence each other, and not all combinations were tested. However, since we have limited resources to run the full parameter optimization process, we use coordinate-wise optimization, which is one of the possible solutions for such problems \citep{friedman2007pathwise}.  

First, we conduct experimenters related to selecting parameters marked with blue color: wavelet type, level of decomposition, and the number of clusters. To compare different parameter sets, we obtain a reference matrix $Y$, which shows whether two intervals are similar ($y_{ij} = 1$) or not ($y_{ij} = 0$) in terms of drilling behavior. Matrix was obtained for the sample of $900$ $t$ segments intervals. For the particular set of model parameters, one could train the KMeans model with $N$ clusters, which as input uses the histogram of labels for $t$ interval, obtained with Bag-of-features approach, and as output gives the label of a cluster for the particular interval. Transforming the vector of predicted labels into a matrix $\hat{Y}$, where each element $\hat{y_{ij}}$ indicates whether the interval $t_i$ and $t_j$ are in the same cluster, one could calculate the Rand Index, showing the similarity between the reference and obtained matrices:

\begin{eqnarray}
    True\ Positive(TP) =\sum_{i} \sum_{j} [y_{ij} = 1] [\hat{y}_{ij} = 1], \\
    False\ Positive(FP) =\sum_{i} \sum_{j} [y_{ij} = 0] [\hat{y}_{ij} = 1] \\
    False\ Negative(FN) =\sum_{i} \sum_{j} [y_{ij} = 1] [\hat{y}_{ij} = 0],\\
    True\ Negative(TN) =\sum_{i} \sum_{j} [y_{ij} = 0] [\hat{y}_{ij} = 0], \\
    Rand\ Index = \frac{TP + TN}{TP + FP + FN + TN}
\end{eqnarray}

Rand Index varies from zero to one, where $Rand\ Index = 1$ means that reference and obtained matrices are identical. Comparing different sets of parameters using such metrics, one could find an optimal parameter setting for the current experiment. In Appendix \ref{Appendix_opt_BOF}, one may see the top $40$ tested sets of parameters with the value of the metric obtained for the optimal $N$. The best sets of parameters include the $Bior\ 2.4$ wavelet and KMeans algorithm with $K = 200$ clusters. To assess the validity of the obtained results, we also calculated the baseline model's metric value, where time-series values were directly used as a feature vector. Rand Index for the baseline approach was $0.840$, which is lower than the best-obtained result, which was $0.878$. 

\begin{figure}[!ht]
    \centering
    \includegraphics[width = 1\linewidth]{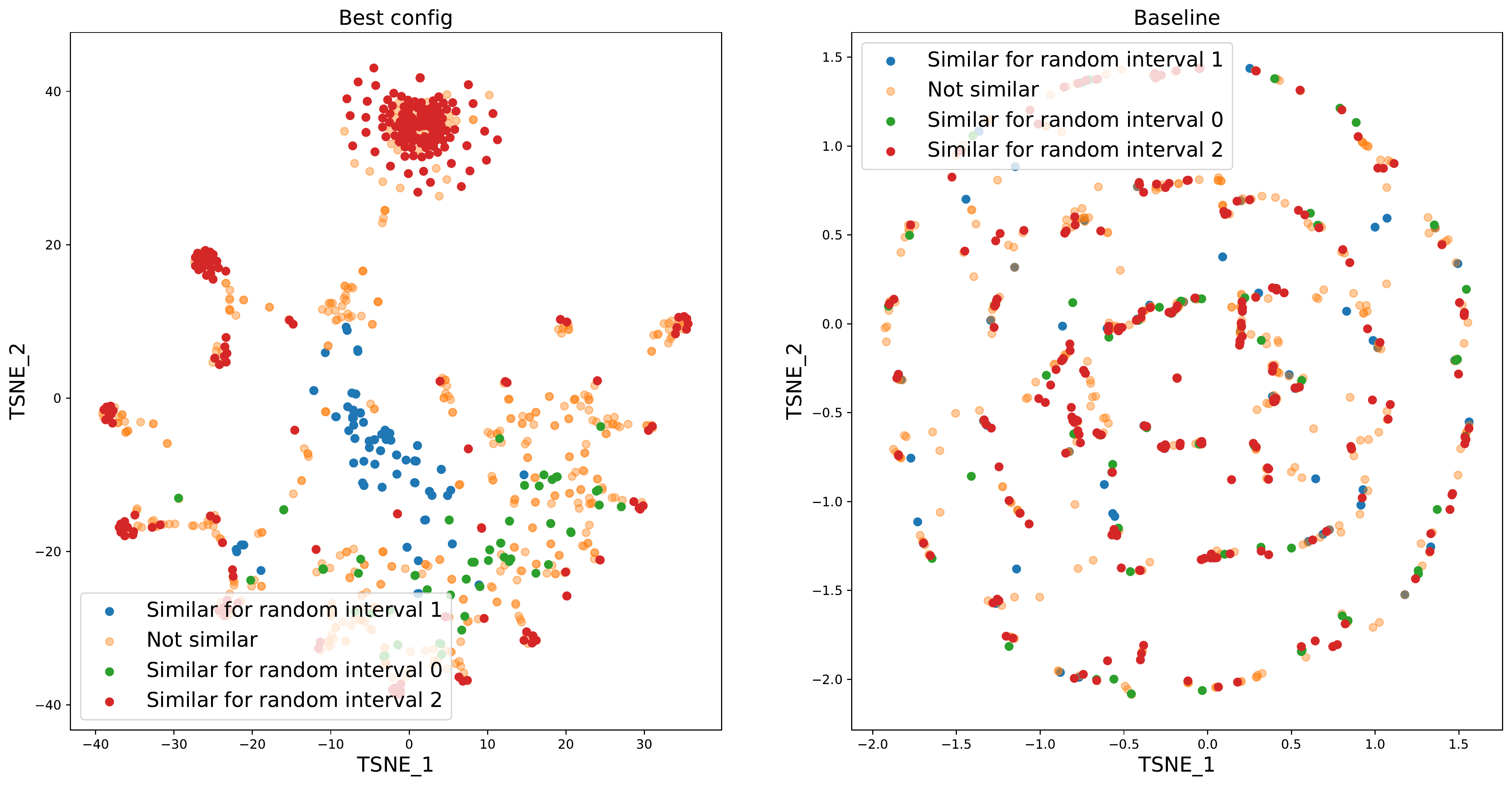}
    \caption{Tsne represents the best-obtained set of parameters and baseline case for the distance matrix between feature vectors of $900$ time series segments. Colors correspond to the group of intervals, similar to randomly picked segments. Bag-of-features representation gives us a better cluster distribution of similar time series segments than the baseline case. } 
    \label{fig:TSNE}
\end{figure}

The results can additionally be validated visually. It can be done using the tsne technique that, as input, use a pairwise distance matrix between features of segments $t$.  Picking randomly several intervals and using their reference matrix values, we could obtain the groups of points corresponding to similar time series segments. The conducted test results for three randomly chosen time series intervals for the best set of parameters and baseline cases are shown in Figure~\ref{fig:TSNE}. Bag-of-features representation gives us a better cluster distribution of similar time series segments than the baseline case. 

\begin{figure}[!ht]
    \centering
    \includegraphics[width = 0.95\linewidth]{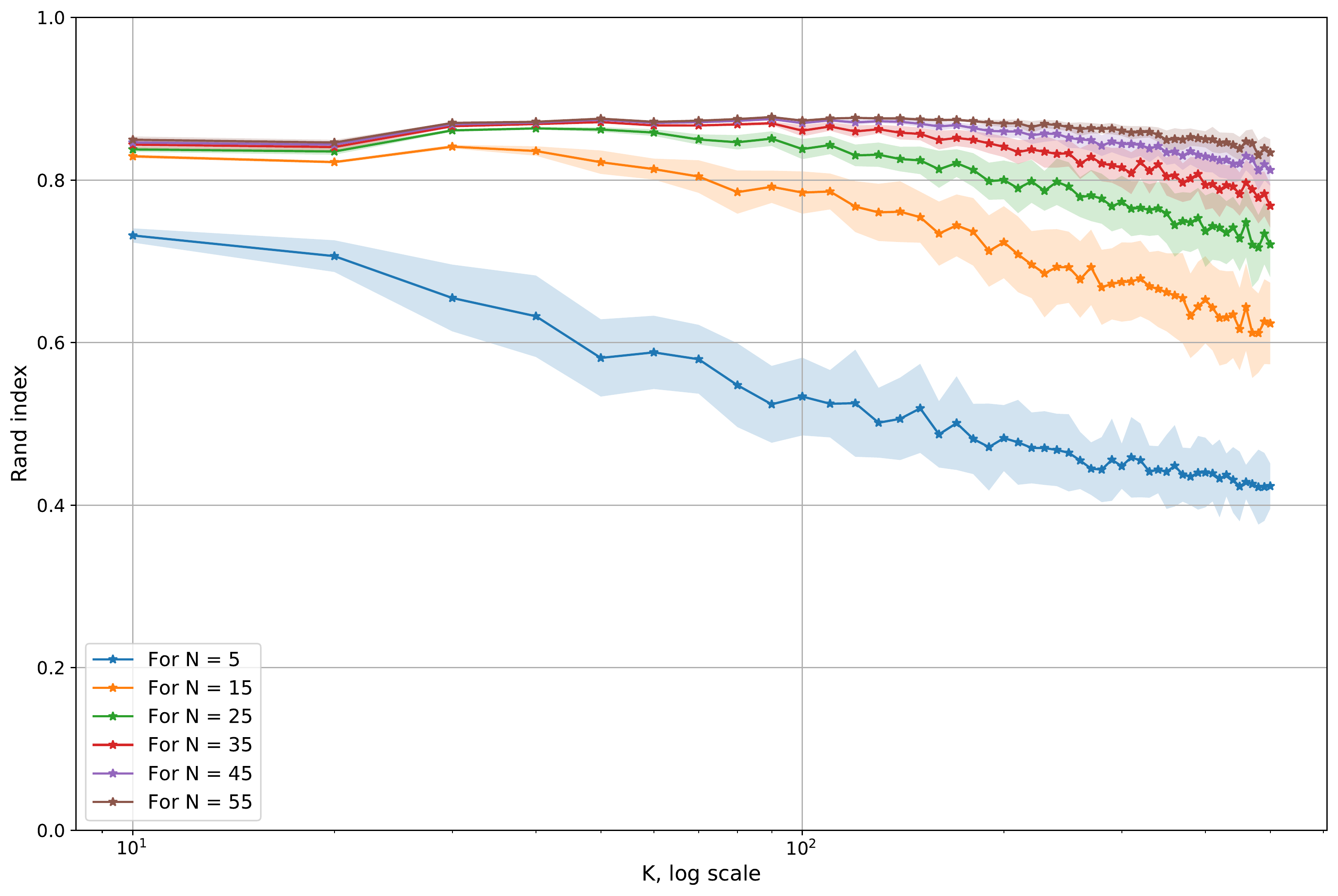}
    \caption{Results of the experiment related to the Bag-of-word approach's sensitivity to the number of clusters in clustering algorithms. The obtained results show that the developed model is sensitive to small values of $N$ (from $0$ to $15$) and $K$ (from $0$ to $40$), while for other values, the Rand Index metric is almost the same.} 
    \label{fig:Sensitivity}
\end{figure}

We also tested the Bag-of-features model's sensitivity to the number of clusters in both clustering algorithms for the best set of parameters. In Figure \ref{fig:Sensitivity}, one may see the Rand Index curves with a confidence interval for different $N$ and $K$ values, where $K$ is represented in the log scale. The obtained results show that the developed model is sensitive to small values of $N$ (from $0$ to $15$) and $K$ (from $0$ to $40$), while for other values, the Rand Index metric is almost the same. Since the best set of parameters was obtained for $N = 25$ and included $K = 200$, the model with the obtained parameters setting can be used in further experiments. 

The second stage of the parameter selection procedure includes selecting the length of segments $t$ and $\tau$ with fixed sets of parameters selected in the first experiment. The selection of these two parameters was carried out based on ROC AUC values obtained during cross-validation. In Table \ref{tab:heatmap} one may see the ROC AUC scores for different sets of the lengths of segments $\tau$ and $t$. If the $\tau$ segment was longer than the length of the $t$ segment, we assumed that the ROC AUC value is undefined. With the $\tau$ segments of length $24$ minutes and $t$ segments of length $72$ minutes model can forecast $70$\% of drilling accidents.

\begin{table}[!ht]
\centering
\caption{ROC AUC scores were obtained on cross-validation for different sets of lengths for segments $t$ and $\tau$ for accident forecasting cases. Using the $\tau$ segments of length $24$ minutes and $t$ segments of length $72$ minutes, the model can forecast 70\% of drilling accidents.}
\resizebox{0.7\textwidth}{!}{%
\begin{tabular}{
>{\columncolor[HTML]{FFFFFF}}c 
>{\columncolor[HTML]{FFFFFF}}c 
>{\columncolor[HTML]{FFFFFF}}c 
>{\columncolor[HTML]{FFFFFF}}c 
>{\columncolor[HTML]{FFFFFF}}c }
\cellcolor[HTML]{FFFFFF} &  & \multicolumn{3}{c}{\cellcolor[HTML]{FFFFFF}t length, minutes} \\ \cline{3-5} 
\multirow{-2}{*}{\cellcolor[HTML]{FFFFFF}} & \multicolumn{1}{c|}{\cellcolor[HTML]{FFFFFF}} & \multicolumn{1}{c|}{\cellcolor[HTML]{FFFFFF}{\color[HTML]{333333} 72 min}} & \multicolumn{1}{c|}{\cellcolor[HTML]{FFFFFF}{\color[HTML]{333333} 180 min}} & \multicolumn{1}{c|}{\cellcolor[HTML]{FFFFFF}{\color[HTML]{333333} 420 min}} \\ \cline{2-5} 
\multicolumn{1}{c|}{\cellcolor[HTML]{FFFFFF}} & \multicolumn{1}{c|}{\cellcolor[HTML]{FFFFFF}8 min} & \multicolumn{1}{c|}{\cellcolor[HTML]{FFFFFF}0.69} & \multicolumn{1}{c|}{\cellcolor[HTML]{FFFFFF}0.7} & \multicolumn{1}{c|}{\cellcolor[HTML]{FFFFFF}0.68} \\ \cline{2-5} 
\multicolumn{1}{c|}{\multirow{-2}{*}{\cellcolor[HTML]{FFFFFF}\begin{tabular}[c]{@{}c@{}}$\tau$ length, \\  minutes\end{tabular}}} & \multicolumn{1}{c|}{\cellcolor[HTML]{FFFFFF}24 min} & \multicolumn{1}{c|}{\cellcolor[HTML]{FFFFFF}0.7} & \multicolumn{1}{c|}{\cellcolor[HTML]{FFFFFF}0.68} & \multicolumn{1}{c|}{\cellcolor[HTML]{FFFFFF}0.67} \\ \cline{2-5} 
\multicolumn{1}{c|}{\cellcolor[HTML]{FFFFFF}} & \multicolumn{1}{c|}{\cellcolor[HTML]{FFFFFF}50 min} & \multicolumn{1}{c|}{\cellcolor[HTML]{FFFFFF}0.68} & \multicolumn{1}{c|}{\cellcolor[HTML]{FFFFFF}0.69} & \multicolumn{1}{c|}{\cellcolor[HTML]{FFFFFF}0.68} \\ \cline{2-5} 
\end{tabular}%
}
\label{tab:heatmap}
\end{table}

\subsection{General model quality}
\label{sec:general_qaulity}
As was mentioned in Section \ref{sec:val_proc}, the general model quality was estimated using the ROC curve and ROC AUC metric.  To benchmark model quality with other models, we trained Convolution Neural Network (CNN), similar to the approaches \citep{aljubran2021deep, qodirov2020development}, which uses $t$ segment values as features, and random classifier, which predicts accident probabilities for each time step in segment $t$ randomly. CNN model has tree convolution units, where each convolution layer is followed by the batch normalization layer, dropout unit (p=0.1), and two fully connected linear layers with $128$ and $32$ input neurons, respectively. For the training, we use SGD optimizer and cross-entropy as loss functions. In addition, to compare the current model with previous studies, we retrained and validated the Breakdown model, presented in paper \citep{antipova2019data}, on the same samples used during the Bag-of-features quality estimation procedures. The breakdown model uses aggregated features, such as mean, slope, etc., over the interval and predicts the accident with the Gradient Boosting algorithm. 

We tuned models' hyperparameters for all models to increase the corresponding metrics values and achieve a fair comparison of the models. For the Bag-of-features and gradient boosting, we adjusted the number of trees, learning rate, and the fraction of observations to be selected for each tree. In contrast, we chose several convolutions layers, learning rate, the maximum number of epochs, and the dropout rate for the CNN model.

\begin{figure}[!ht]
\includegraphics[width = 0.75\linewidth]{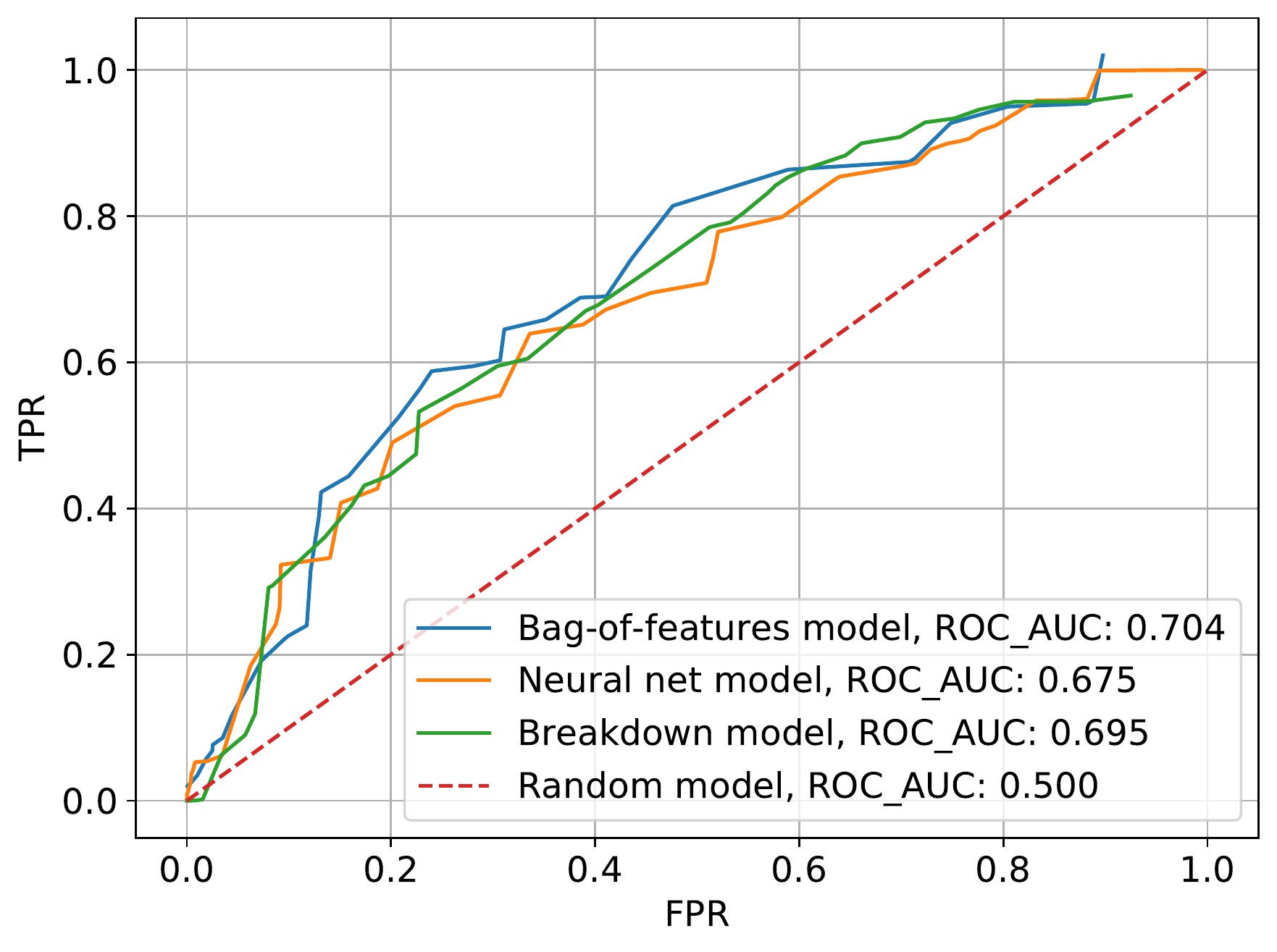} 
\caption{Accident forecasting quality metrics for the Bag-of-features, Convolution Neural Net, Breakdown, and Random models. As one may notice, the Bag-of-features model gives us a higher value of the ROC AUC metric than other models.}
\label{fig:metrics}
\end{figure}

According to the model quality graph shown in Figure \ref{fig:metrics}, the ROC AUC score for the Bag-of-features model is $0.704$,  while for the CNN, Random models, the ROC AUC score is $0.675$ and $0.5$, respectively. Comparing the Bag-of-features model with the Breakdown model, the current model gives us a higher metric value. 

\begin{figure}[!ht]
\includegraphics[width = 0.75\linewidth]{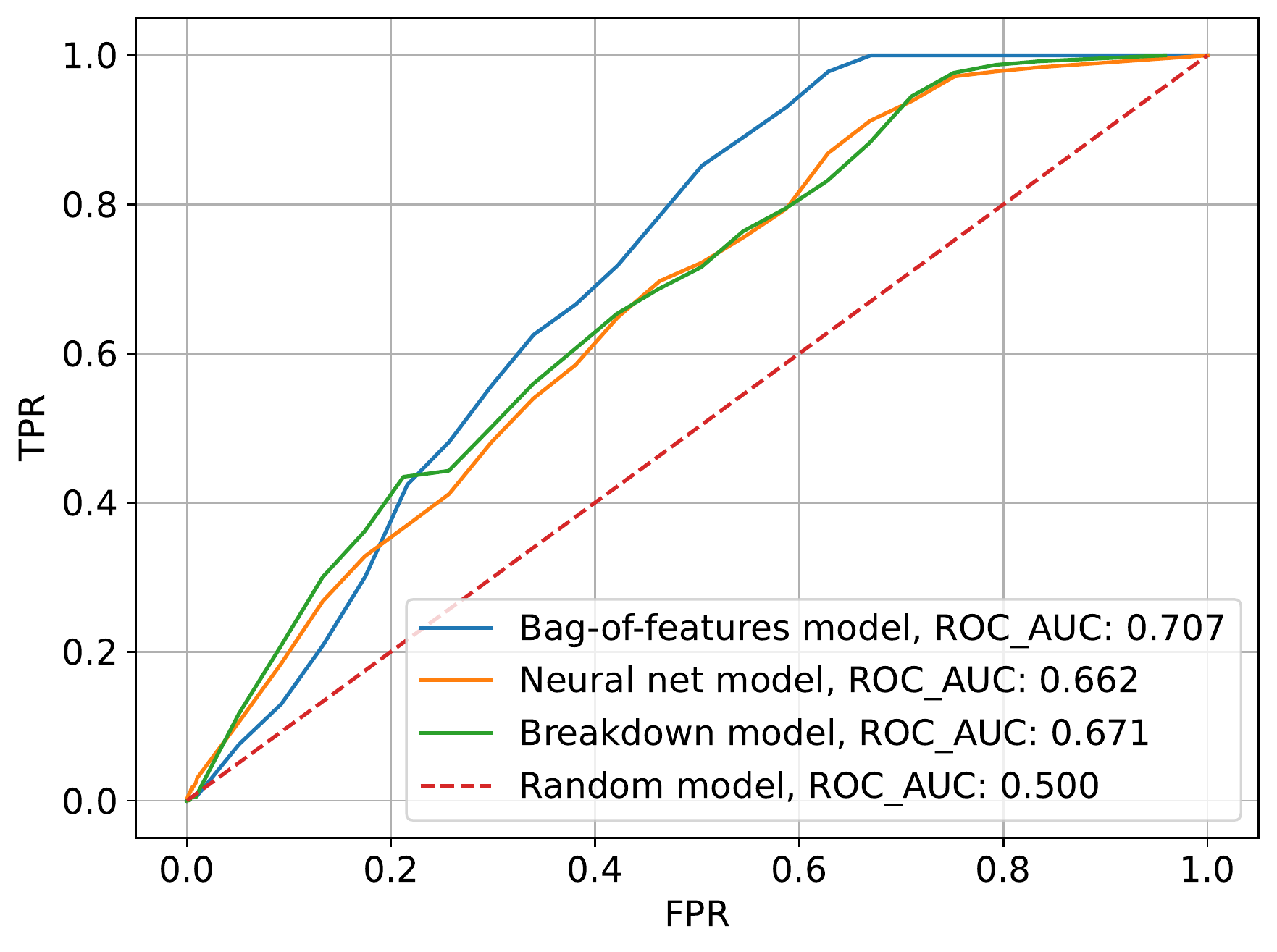} 
\caption{Accident forecasting quality metrics for the Bag-of-features, Convolution Neural Net, Breakdown, and Random models for the binary classification problem. As one may notice, the Bag-of-features model gives us a higher value of the ROC AUC metric than other models.}
\label{fig:metrics_binary}
\end{figure}

Additionally, we validate model quality without considering multiclass classification, but only considering if there is an accident or not at a particular time moment. On Figure \ref{fig:metrics_binary} one may see the corresponding ROC AUC values. Similar to the previous case, the Bag-of-features model performs better than other models. 

Comparing the Bag-of-features performance for multiclass and binary problems, one may notice that the difference between ROC AUC metric values is insignificant. It means that the model clearly distinguishes the pre-accident situation of a different type, and most of the errors relate to the misinterpretation of whether the segment contains a pre-accident pattern or not.

\begin{figure}[!ht]
    \centering
    \includegraphics[width = 1\linewidth]{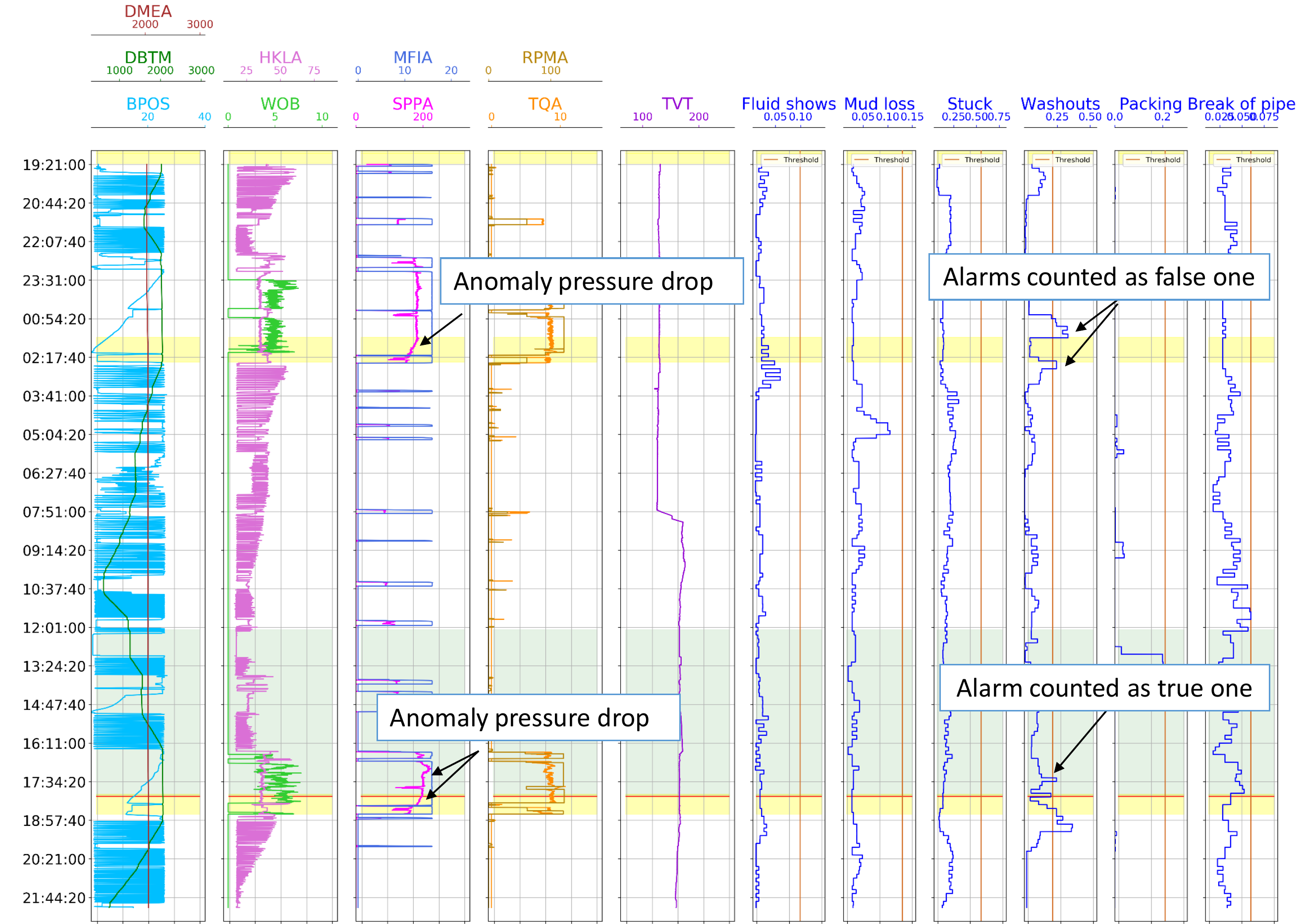}
    \caption{Mud log with forecasted probabilities of accidents. The yellow area stands for the anomaly behavior associated with anomaly pressure drop. However, such anomaly behavior did not lead to the washout accident in this case, and model alarms were counted as false ones. The green area corresponds to the six hours interval before another washout accident. The model alarm that happened in this area was counted as a true one. In both cases, the model was triggered on anomaly pressure drops; however, according to the metric calculation scheme, only the last alarms were considered as the true ones.} 
    \label{fig:prob_example}
\end{figure}

According to Figure \ref{fig:metrics}, the Bag-of-features model can forecast 70\% of drilling accidents with a false positive rate equals to $40$\%.  It corresponds to the four false alarms per day for one hour long. The high value of the false-positive rate for the model can be explained by the fact that the testing sample also contains anomaly behavior that did not lead to the accident, however, the model triggered that. An example of such a situation is shown in Figure \ref{fig:prob_example}. The yellow area stands for the anomaly behavior associated with an anomaly pressure drop that did not lead to the actual washout accident.  In contrast, in the green zone, which corresponds to the six hours interval before another washout accident, one may observe similar pressure reduction, which leads to the washout accident in this case. The model alarmed the right types during both anomaly behavior intervals. However, according to the metric calculation scheme, the first two alarms were count as false ones. An additional expert evaluation of the mud logs, where such intervals will be highlighted as abnormal regions, is required to take into account such cases.

The current metric point corresponds to 70\% of forecasted drilling accidents with a false positive rate equal to $40$\%, achieved with a false negative rate (FNR) equal to 30\%. In Table \ref{tab:accident_types_non_predicted} one could see the number of non-predicted cases for each accident type. The model is struggled with breaks and washouts of drilling pipe cases, while other accidents showed good results. It could be explained that the break of drilling pipe usually does not have any anomaly behavior before the accident. However, a detailed evaluation of such cases is necessary.

\begin{table}[!ht]
\centering
\caption{The number of non-predicted drilling accidents by type for the particular threshold, corresponding to the TPR, equals 70\%, and FPR equals 40\%. }
\begin{tabular}{|c|c|c|c|}
\hline
 & Accident type & Number of non-predicted cases & Total number of cases \\ \hline
1 & Stuck & 5 & 55 \\ \hline
2 & Washouts of drilling pipe & 8 & 20 \\ \hline
3 & Mud loss & 5 & 16 \\ \hline
4 & Breaks of drilling pipe & 12 & 15 \\ \hline
5 & Fluid shows & 4 & 10 \\ \hline
6 & Packing & 3 & 9 \\ \hline
\end{tabular}
\label{tab:accident_types_non_predicted}
\end{table}

\subsection{Optimisation of the model performance}
\begin{figure}[!ht]
    \centering
    \includegraphics[width = 0.75\linewidth]{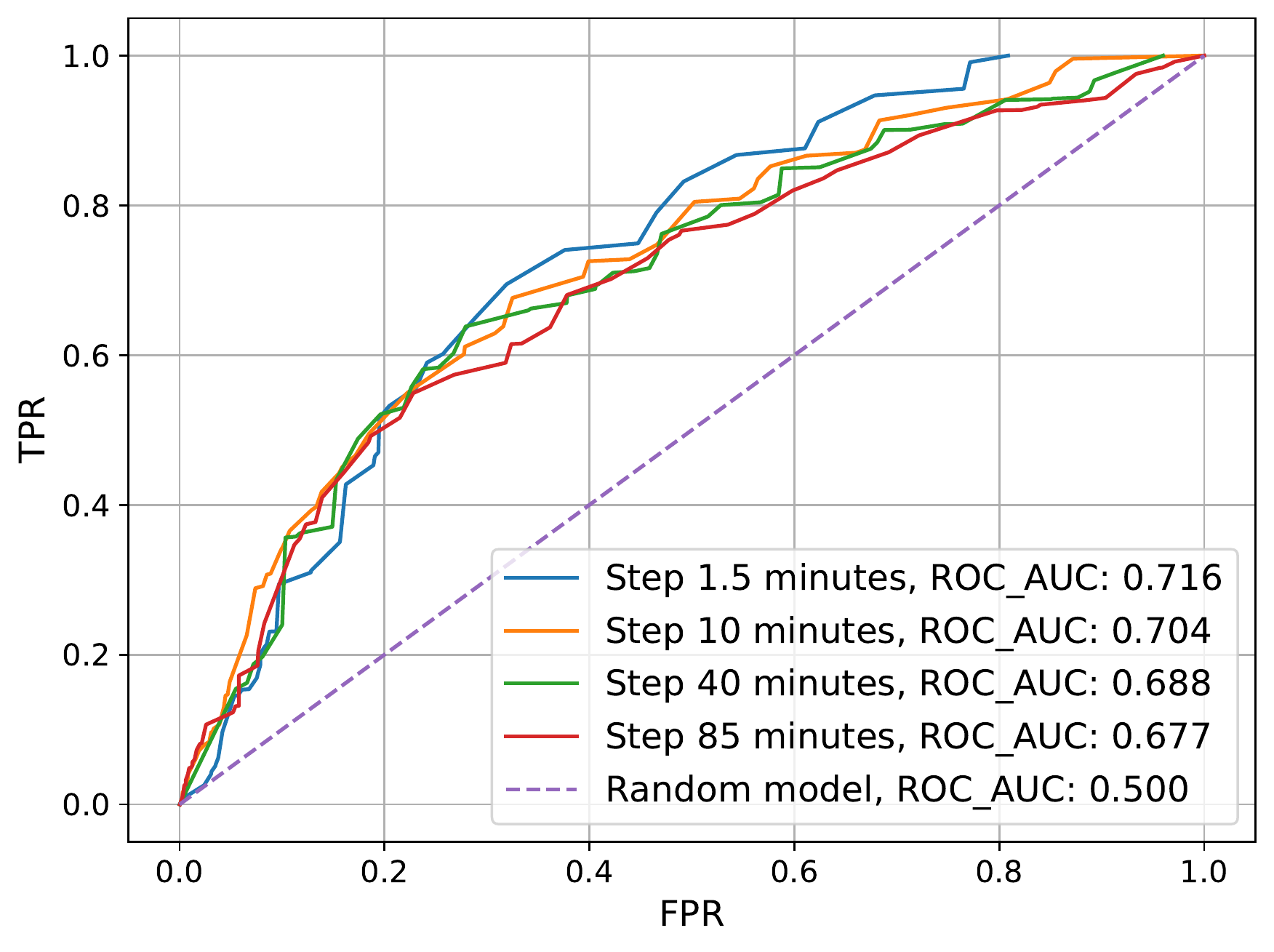}
    \caption{Model quality metrics were obtained for different steps for $t$ segments generation. Using steps for $t$ segments generation less than one hour long does not significantly affect the model's quality.} 
    \label{fig:steps}
\end{figure}

The last experiment selected the frequency with which the model should be turned on during real-time testing. Fixing the parameters selected during previous stages, we obtained the metric values for different steps for $t$ segments generation (Figure \ref{fig:steps}). One may notice that using steps for $t$ segments generation less than one hour long does not significantly affect the model's quality. Since there is still a probability of missing the non-standard behavior with more than $20$ minutes, we decided to use step equals $10$ minutes for further production testing.

\subsection{Discussion and future work}
Forecasting accidents during well drilling is essential for the oil and gas industry since it would help save time and money related to the accident consequences. Nowadays, it becomes evident that a combination of drilling engineers' knowledge and machine learning models allows us to predict accidents of several selected types. 

In a separate section, we conduct an ablation study of the model's main hyperparameters. Conducted tests suggest that the current model is not sensitive to the changes in parameter setting. The Bag-of-features model was also compared to the other state-of-the-art models such as the CNN model, the Breakdown model, described in paper \citep{antipova2019data}, and the Random model. The developed model's quality metrics exceed the quality of other models and ensure good performance during real-time drilling operations.

However, there is still room for improvement. In particular, it is interesting to conduct experiments to identify the dependence of model quality on the size of the accident database and obtain model quality, using other metrics that might indicate the ratio of true forecasted events and the particular number of false alarms per day.  Besides, as mentioned in section \ref{sec:general_qaulity}, additional expert evaluation of the mud logs is required to take into account anomaly drilling behavior that happened outside the six-hour interval before the accident. It also might be a good strategy for the objective function during model training to assign higher penalties for false negatives of more severe drilling accidents.

Another problematic aspect of this method is the highly unbalanced classification problem since most data patterns are rare. Moreover, the patterns are fuzzy and can be distorted both in time and in the time series' values by noise and scale. Thus, it is necessary to test and find another time series representation where such patterns might be better observed.

Nowadays model is testing in real oilfields in Russia. To operate the model, we developed software integrated with the Wellsite Information Transfer Standard Markup Language (WITSML) data server into clients' existing IT infrastructure. All calculations take place on the cloud and, therefore, do not require significant additional computing power.

\section{Conclusions}
\label{sec:conclusion}
We have demonstrated that the model is generalizable from field to field. The scalability illustrates that the model tackles the key challenge of drilling accident forecasting: the lack of properly defined pre-accident patterns for different drilling regimes at different fields. 

The novelty of the research is the Bag-of-features representation of multivariate telemetry logs and the machine learning model itself. Talking numbers, the developed model can forecast 70\% of the accidents with a false positive rate equal to $40$\%, which is higher than the metrics of the state-of-the-art studies and random baseline. Moreover, obtained results show that model clearly distinguishes the pre-accident situation of a different type, and most of the errors relate to the misinterpretation of whether the segment contains a pre-accident pattern or not. 

The intelligent system for real-time drilling accident prediction allows preventing drilling accidents and reducing costs associated with failures. In the future, we plan to develop the model further for pushing down the false positive rate and increasing the timing of correct predictions. For these purposes, we will work thoroughly with the training set marking the abnormal cases which are not followed by accident and developing the physics-informed filters on top of the core model.

\bibliographystyle{spbasic}      
\bibliography{BOF}   

\begin{appendices}
\newpage
\section{}
\label{Appendix_example}
\begin{figure}[!ht]
    \centering
    \includegraphics[width = 0.85\linewidth]{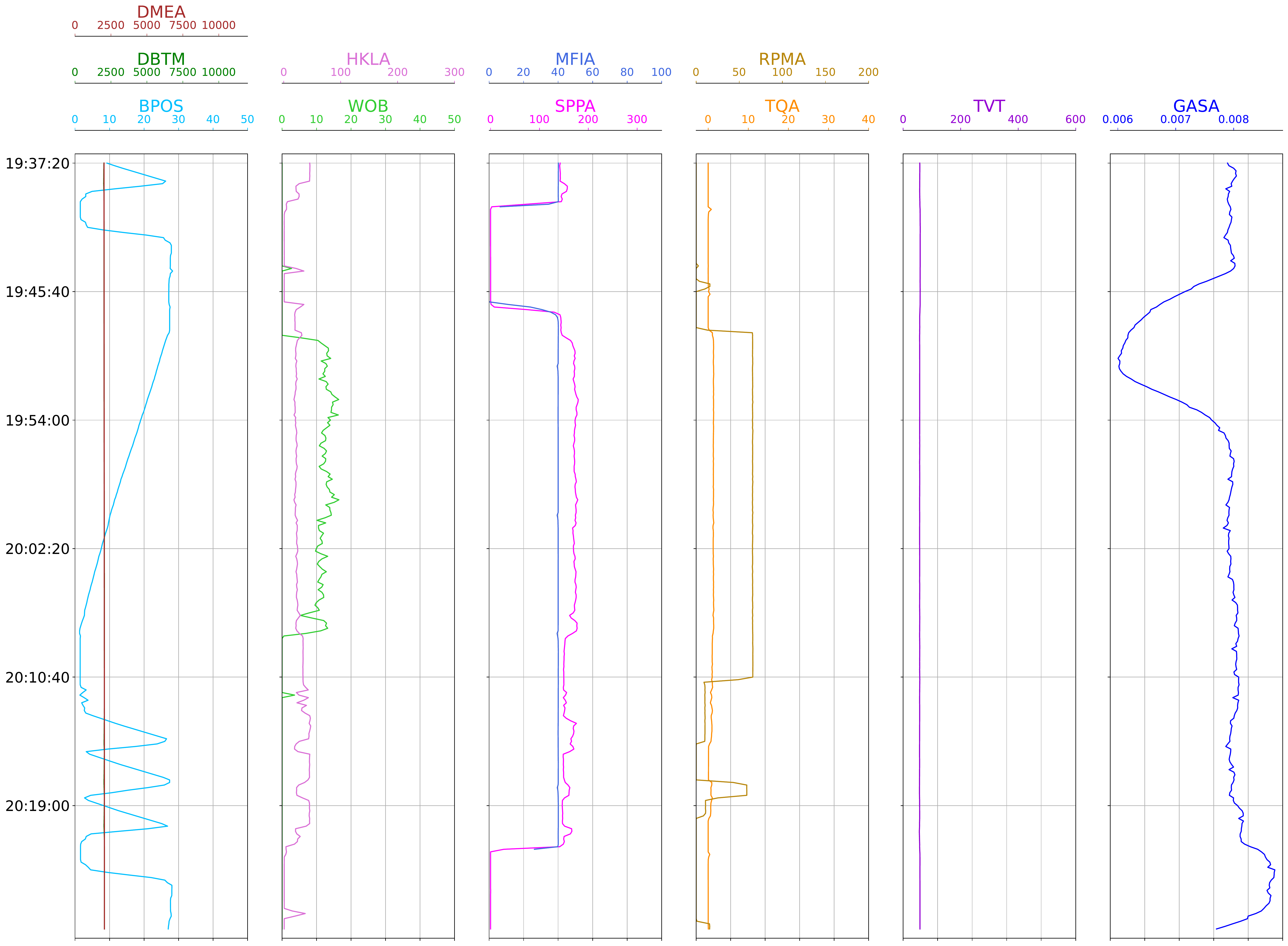}
    \caption{Example of mud telemetry data that was used for feature generation procedure.} 
    \label{fig:MWD_exp}
\end{figure}

\begin{figure}[!ht]
    \centering
    \includegraphics[width = 0.85\linewidth]{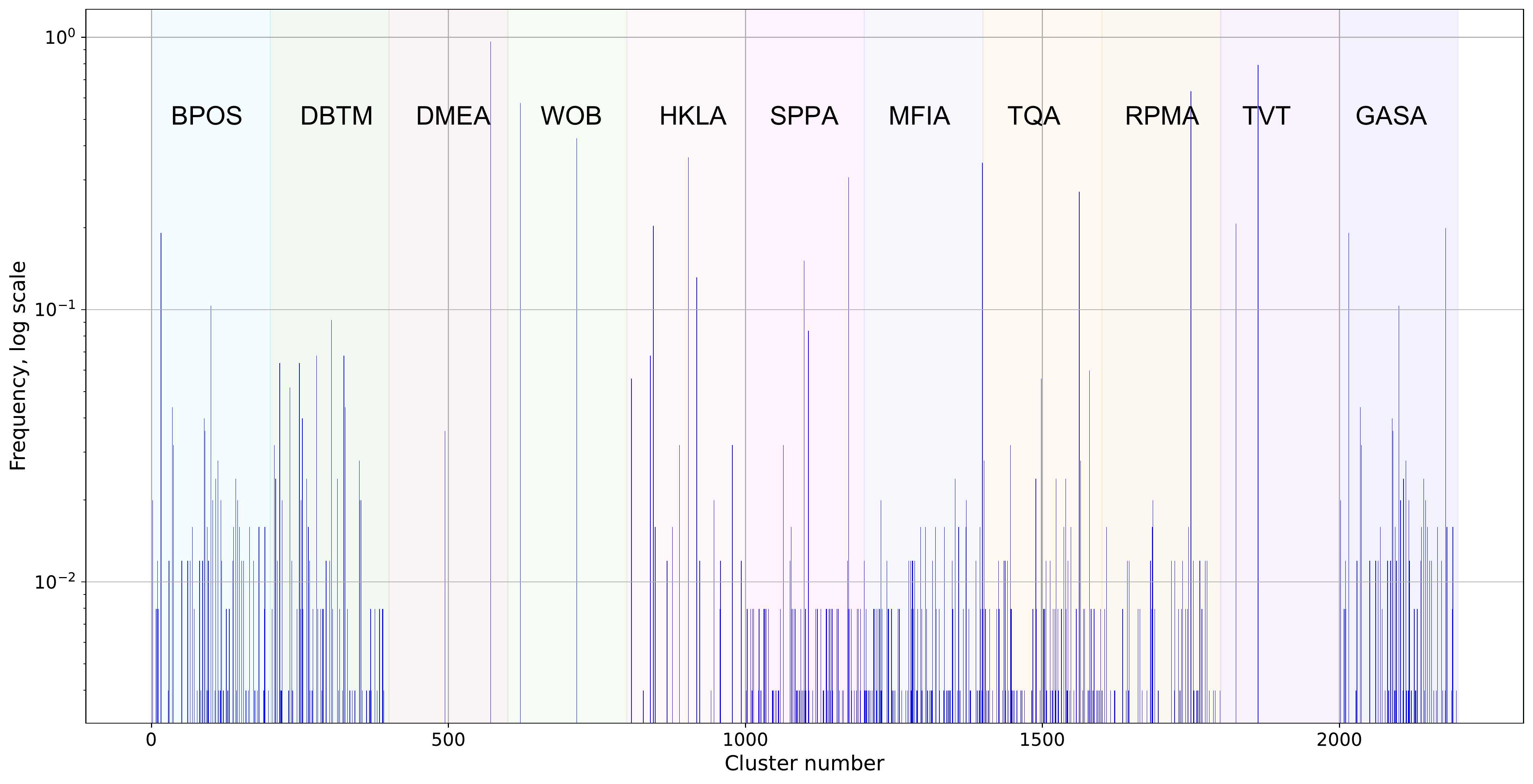}
    \caption{Features obtained from the mud telemetry data on fig \ref{fig:MWD_exp}, that used as input for the classification model.} 
    \label{fig:Hist_exp}
\end{figure}

In Figure \ref{fig:MWD_exp} one may see the example of mud telemetry data used during the drilling accident forecasting problem. After the feature generation procedure, the presented part of the time-series is transformed into a histogram, presented in Figure \ref{fig:Hist_exp}.

\newpage
\section{}
\label{Appendix_opt_BOF}
Table \ref{tab:opt_BOF}  shows the results of the first stage of an optimization procedure for the top $40$ parameter sets.  
\begin{table}[H]
\centering
\caption{Top $40$ parameters sets that were tested during the first optimization stage of the Bag-of-features approach. Rand Index for baseline approach was 0.84, which is lower than the best-obtained result (0.87).}
\resizebox{0.75\textwidth}{!}{%
\begin{tabular}{|c|c|c|c|c|c|}
\hline
 & K & level & Wavelet & Rand Index & N \\ \hline
1 & 200 & 3 & bior2.4 & 0.878484 & 25 \\ \hline
2 & 100 & 5 & bior2.4 & 0.877657 & 25 \\ \hline
3 & 100 & 4 & bior2.4 & 0.877504 & 25 \\ \hline
4 & 100 & 4 & bior2.4 & 0.877341 & 25 \\ \hline
5 & 100 & 4 & coif5 & 0.877247 & 25 \\ \hline
6 & 200 & 4 & db3 & 0.877114 & 25 \\ \hline
7 & 100 & 3 & coif5 & 0.876965 & 25 \\ \hline
8 & 100 & 5 & coif5 & 0.87681 & 25 \\ \hline
9 & 100 & 5 & db3 & 0.876686 & 25 \\ \hline
10 & 100 & 4 & coif5 & 0.876612 & 25 \\ \hline
11 & 200 & 3 & coif5 & 0.876277 & 25 \\ \hline
12 & 100 & 3 & coif5 & 0.8762 & 25 \\ \hline
13 & 100 & 3 & bior2.4 & 0.875928 & 22 \\ \hline
14 & 100 & 5 & coif5 & 0.875765 & 22 \\ \hline
15 & 200 & 3 & db3 & 0.875573 & 25 \\ \hline
16 & 100 & 4 & bior2.4 & 0.875506 & 25 \\ \hline
17 & 100 & 5 & bior2.4 & 0.875365 & 25 \\ \hline
18 & 200 & 3 & bior2.4 & 0.875281 & 25 \\ \hline
19 & 100 & 3 & bior2.4 & 0.875178 & 25 \\ \hline
20 & 100 & 4 & coif5 & 0.875136 & 22 \\ \hline
21 & 100 & 3 & coif5 & 0.875064 & 25 \\ \hline
22 & 100 & 4 & db3 & 0.874835 & 25 \\ \hline
23 & 100 & 5 & bior2.4 & 0.874827 & 22 \\ \hline
24 & 200 & 4 & bior2.4 & 0.87479 & 25 \\ \hline
25 & 200 & 3 & coif5 & 0.87478 & 25 \\ \hline
26 & 100 & 3 & bior2.4 & 0.874499 & 25 \\ \hline
27 & 100 & 5 & db3 & 0.874368 & 25 \\ \hline
28 & 200 & 5 & coif5 & 0.874316 & 25 \\ \hline
29 & 100 & 5 & coif5 & 0.874158 & 25 \\ \hline
30 & 200 & 3 & db3 & 0.874094 & 25 \\ \hline
31 & 200 & 3 & coif5 & 0.874079 & 25 \\ \hline
32 & 200 & 4 & db3 & 0.874035 & 25 \\ \hline
33 & 100 & 4 & db3 & 0.87396 & 25 \\ \hline
34 & 100 & 4 & db3 & 0.87396 & 25 \\ \hline
35 & 200 & 5 & bior2.4 & 0.873684 & 25 \\ \hline
36 & 200 & 5 & bior2.4 & 0.87363 & 25 \\ \hline
37 & 100 & 3 & db3 & 0.873504 & 25 \\ \hline
38 & 200 & 3 & bior2.4 & 0.873353 & 25 \\ \hline
39 & 200 & 5 & coif5 & 0.873131 & 25 \\ \hline
Baseline & Baseline & Baseline & Baseline & 0.840027 & 45 \\ \hline
\end{tabular}}

\label{tab:opt_BOF}
\end{table}

\end{appendices}
\end{document}